# Abstractive Text Summarization for Resumes With Cutting Edge NLP Transformers and LSTM

Öykü Berfin Mercan, Sena Nur Cavsak, Aysu Deliahmetoglu (Intern), Senem Tanberk
oyku.berfin.mercan@huawei.com, sena.nur.cavsak@huawei.com, aysudeliahmetoglu@gmail.com, 0000-0003-1668-0365
Huawei Turkey Research and Development Center, Istanbul

*Abstract*— Text summarization is a fundamental task in natural language processing that aims to condense large amounts of textual information into concise and coherent summaries. With the exponential growth of content and the need to extract key information efficiently, text summarization has gained significant attention in recent years. In this study, LSTM and pre-trained T5, Pegasus, BART and BART-Large model performances were evaluated on the open source dataset (Xsum, CNN/Daily Mail, Amazon Fine Food Review and News Summary) and the prepared resume dataset. This resume dataset consists of many information such as language, education, experience, personal information, skills, and this data includes 75 resumes. The primary objective of this research was to classify resume text. Various techniques such as LSTM, pre-trained models, and fine-tuned models were assessed using a dataset of resumes. The BART-Large model fine-tuned with the resume dataset gave the best performance.

## I. Introduction

In the recent year, Natural Language Processing (NLP) becomes a popular research area. With the help of advances in technology, a large number of information and documents are collected in the form of text data in the digital world rapidly. In order to obtain meaningful results from text data efficiently, researchers study to develop NLP tasks such as text classification, question answering, text generation and text summarization. The task of text summarization in NLP has become a research of interest, considering the rapid increase in the number of documents, it is important to minimize the time wasted and unnecessary information density in the process of obtaining useful information from the document. Text summarization aims to create, short and accurate summary from the large text data without human intervention. Text summarization is separated into two main types; extractive text summarization and abstractive text summarization. Extractive text summarization creates summary with selected important information using the same words from main text. Differently, abstractive text summarization creates better summary with different words and flexible representations as humans. With the advance in deep learning, many studies have focused on abstractive text summarization [1, 2]. Song et al [1] developed an LSTM-CNN-based Abstracting Text Summarization model. Firstly, they extracted the sentences from the source sentences with the Multiple Order Semantic Parsing model. Then they created text summaries using the deep learning method. The authors used CNN/Daily Mail and Gigaword data for this model and compared performance. Hanunggul et al. [2] examined the effect of local attention in the LSTM model to generate abstract text summaries. They used the Amazon Fine Food Review dataset and evaluated the performance of the model using the GloVe dataset. The findings showed that the ROUGE-1 outperformed the global attention-based model as it produced more words in the actual summary. On the other hand, the local attention-based model achieved higher ROUGE-2 scores because it generated more word pairs found in the actual summary.

Abstractive text summarization gives succeed performance with transformer architecture-based pre-trained language models [3, 4, 5, 6, 7, 8, 9, 10]. Zolotareva et al. [3] used Sequence-to-Sequence Recurrent Neural Networks and Transfer Learning techniques with Composite Text-to-Text Converter for the text summarization problem. They developed the Transfer Learning-based model for Abstractive text summarization. They used the Transformer or T5 framework from the BBC News dataset. Ranganathan and Abuka [4] introduced a text summarization method based on the converter architecture, specifically the Text-to-Text Converter (T5) model. The goal was to condense long texts into concise but informative summaries. The researchers did this for the Irvine (UCI) drug reviews dataset, by training and testing the T5 model on human-generated summaries. In addition, PEGASUS improvements were made using the T5 model BBC News dataset. Zhang et al [5] offered the model for abstractive text summarization. The researchers explored different methods for selecting gap sentences and found that choosing principle sentences yielded the best results. By optimizing the model's configuration, they achieved state-of-the-art performance on 12 datasets. Lalitha et al [6] used various abstractive summarization techniques, including T5, BART, and PEGASUS. These techniques aimed to extract essential information from medical documents and to provide concise summaries suitable for users' interests. They used ROUGE metrics to evaluate the performance of these models. Among the tested models, the most effective model was the PEGASUS model with a ROUGE score of 0.37. In [7], Borah et al. evaluated abstractive text summarization performance of T5 on open-source datasets which are CNN/Daily Mail, MSMO and XSUM. It showed that T5 gives short and fluent summary and the best results obtained from MSMO dataset. Another study [8] compared abstractive text summarization performance of pre-trained models which are BART, T5 and PEGASUS and BBC News Dataset. Pre-trained models from HuggingFace were finetuned and evaluated for summarization. Experiment showed that the T5 model gives the highest ROUGE score. Another similar study is [9],

Yadav et al. proposed BART model that was finetuned with the Amazon Fine Food Review dataset. Model was compared with previous studies in the literature, it was seen that a successful result was obtained. In [10] Rehman et al. analyzed abstractive text summarization using different models and datasets. BART and PEGASUS models had the highest ROUGE on CNN/DailyMail and SAMSum datasets, while PEGASUS showed accurate performance on BillSum. BART and PEGASUS got better result than T5.

In the literature review, it was observed that abstractive text summarization was carried out with many open-source or custom datasets, such as news, academic review and product review, and conversation. However, to the best of our knowledge, the study has not been published on abstractive resume text summarization using pre-trained language models. In this study was focused on abstractive resume text summarization. Resume dataset that includes language, education, experience, personal information, and skills information was created. LSTM model was trained and BART, T5, PEGASUS were finetuned using resume dataset. Models performance is evaluated and BART-Large model gave the highest ROUGE score. In addition, open-source datasets were used for training and finetuning. Performance of models were evaluated and compared to each other (Fig1).

## II. METHOD

### A. Dataset

*a) Xsum- Extreme Summarization Dataset*

The XSum dataset has been crafted for the training needs of language models focusing on extreme summarization. It consists of pairs of summary-article, with over 200,000 news articles written in English. Each article is 500 to 800 words long, while the corresponding summaries are approximately 10 to 30 words [11].

*b) CNN/Daily Mail*

The CNN/Daily Mail dataset is a widely utilized resource in the field of natural language processing and machine learning, particularly for text summarization tasks. It includes a collection of news articles sourced from both CNN/Daily Mail, accompanied by corresponding multi-sentence summaries. The dataset covers a variety of topics and consists of longer articles and shorter summaries that capture the main points [12].

*c) Amazon Fine Food Review*

The Amazon food reviews dataset is a collection of customer feedback and ratings for food products available on Amazon. Customers rate the purchased food items on a scale of 1 to 5, reflecting their satisfaction. The dataset includes the text of customer reviews, which offers qualitative insights into their experiences and sentiments. The dataset undergoes preprocessing to ensure its quality and consistency by removing duplicates, handling missing values, and filtering irrelevant information [13].

*d) News Summary*

The dataset comprises 4,515 examples and includes the following information: Author_name, Headlines, Url of Article, Short text, and Complete Article. The data was collected by gathering summarized news in shorts and scraping news articles from The Hindu, Indian Times, and The Guardian.

*e) Resume*

We have a dataset consisting of 280 resumes, for which we have performed the necessary anonymization and labeling processes. The resumes have been divided into five sections: language, education, experience, personal information, and skills. In this dataset, we specifically focused on the experience section and obtained 75 resumes containing relevant information. The dataset can be utilized for various purposes such as language education, recruitment processes, talent evaluation, and more. The experience section provides insights into candidates' past work experiences, including job roles, responsibilities, and company details. By implementing anonymization and labeling techniques, we have ensured privacy and data protection by safeguarding personal information.

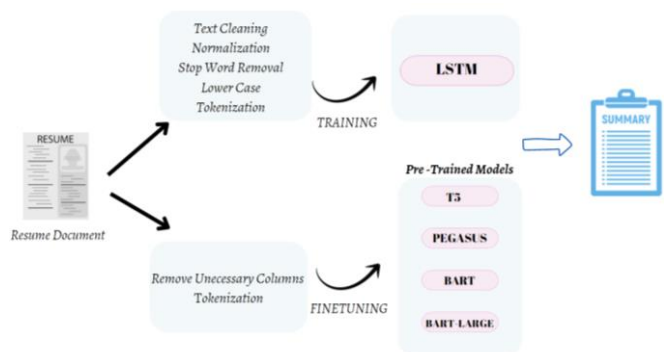

*Fig 1.Overview*

### B. Preprocessing

Preprocessing plays a crucial role in language processing models like LSTM for text summarization. By performing preprocessing steps, we enhance the model's ability to generate consistent, focused, and meaningful summaries. These preprocessing steps encompass tasks such as text cleaning, normalization, stop word removal, and tokenization.

- To ensure the model disregards irrelevant information in the text, unnecessary characters, special symbols, numbers, and punctuation marks are eliminated from the text.
- To enhance the consistency of the model's vocabulary, all words are converted to lowercase, disregarding the distinction between uppercase and lowercase letters in the text.
- Grammatically insignificant words have minimal impact on the text's overall meaning. Consequently, stop words are removed from the text during the summarization process, facilitating the model in creating more focused and concise summaries.

Models such as LSTM require the text to be broken down into smaller units, known as tokens. This process involves subdividing the text into meaningful units such as words, sentences, or even sub-sentences, enabling the model to operate at the word level and perform more fine-grained processing and summary generation.

*C. Models*

   *a) LSTM* (Long Short-Term Memory)

A popular recurrent neural network (RNN) architecture for deep learning is called LSTM. RNN networks and LSTM networks have a lot in common. An artificial neural network called LSTM was created expressly to overcome the problems with RNNs. Traditional RNNs have an issue with vanishing gradients, which can lead to the loss of critical information from previous steps and information loss over lengthy sequences. LSTM networks resolve this problem by incorporating gates that manage and control the information flow. With the aid of these gates, LSTM networks can manage long-term dependencies and the issue of disappearing gradients with greater effectiveness [14].

   *b) BART* (Bidirectional and Auto-Regressive Transformers)

BART is a powerful translation and summarization model used in the field of natural language processing. BART is a Transformer-based model that exhibits outstanding performance in language tasks. Its combination of bidirectional and auto-regressive properties makes it effective for both translation and summarization tasks. With its bidirectional capability, it excels in understanding texts and capturing context. The auto-regressive feature allows it to generate fluent and coherent summaries based on the original text. The BART model is trained by pretraining on a large dataset and then fine-tuning on task-specific data. As a result, it can achieve excellent results in translation and summarization domains [15].

   *c) Bart-Large*

BART-Large is a variant of the BART (Bidirectional and Auto-Regressive Transformers) model, which is a powerful language generation model. BART-Large is specifically trained on a large-scale dataset and has a larger model size compared to the base BART model. With increased capacity and parameters, BART-Large demonstrates enhanced performance in various natural language processing tasks such as text summarization, translation, and text generation. The larger size of BART-Large allows it to capture more intricate language patterns, leading to improved language understanding and generation capabilities [15].

   *d) Pegasus-X*

Pegasus-X is an abstract text summarization model built on the GPT-3 language model. In order to learn the general structure of the language and to improve the ability to understand and summarize abstract texts, it goes through two stages: preliminary training and fine-tuning. This model performs impressively in transforming original texts into concise, coherent and meaningful summaries [16].

   *e) T5 (Text-To-Text Transfer Transformer)*

T5 is a powerful language model developed by Google Research that revolutionizes natural language processing tasks. Unlike traditional models designed for specific tasks, the T5 is a unified model capable of performing a wide variety of text-based tasks, including text classification, answering questions, summarizing and translating. The T5's approach is based on a "text-to-text" framework, where it converts all tasks into a text-to-text format that makes it easy to tackle different tasks with a single model. The T5 achieves remarkable results by pre-training on a large-scale dataset and fine-tuning the task-specific data. [17].

III. EXPERIMENTS

In chapter 2, datasets, preprocessing and models are explained. In this section, the performance of the models were analyzed. The text summarization performance of the LSTM with the pre-trained models that are BART, T5, and Pegasus were compared to each other. Then models were analyzed for resume text summarization. In the study, open-source data sets, which are widely used in text summary studies, were used as train, validation and test data. Models were trained and finetuned on four open source datasets which are Xsum, Cnn/Daily Mail, News Summary, Amazon Fine Food Review and our dataset which is called the resume dataset. The performance of the models were evaluated using ROUGE (Recall-Oriented Understudy for Gisting Evaluation) metric. ROUGE is one of the most popular metrics to evaluate model for text summarization. Metrics compare the similarity between reference summary and summary of model, and then produce accuracy according to the similarity. ROUGE analyzes the performance of the text summarization model using N-gram. ROUGE_N presents an overlap of N-gram between reference and model output. ROUGE_1 score is referred based on the overlap of each word (unigrams) between the reference text and the model output. ROUGE_2 is referred to base on the overlap of bigrams between the reference text and the model output. ROUGE_L is computed based on the longest common subsequences. It compares reference text and model output based on sentence-level similarity using the longest common subsequences. The summarization performance of each model was evaluated using the ROUGE, which is defined by recall (R) (1, 3), precision (P) (2, 4), and f-measure (F) (5) scores.

$$Recall = \frac{\# \ of \ n-grams \ overlapping \ words}{\# \ of \ n-gram \ in \ the \ reference \ summary} \quad (1)$$

$$Precision = \frac{\# \ of \ n-grams \ overlapping \ words}{\# \ of \ n-gram \ in \ the \ summary \ of \ the \ model} \quad (2)$$

These metrics are updated for ROUGE_L;

$$Recall = \frac{\# \ of \ words \ in \ longest \ common \ subsequence}{\# \ of \ n-gram \ in \ the \ reference \ summary} \quad (3)$$

$$Precision = \frac{\# \ of \ words \ in \ longest \ common \ subsequence}{\# \ of \ n-gram \ in \ the \ summary \ of \ the \ model} \quad (4)$$

$$F - mesaure = 2 * \frac{Precision * Recall}{Precision + Recall} \quad (5)$$

Recall is computed ratio by the ratio of the number of n-grams overlapping words by the total number of n-grams in the summary of the model. Precision is computed by the ratio of the number of n-grams overlapping words by the total number of n-grams in the reference summary. The metric score is between 0 and 1, with 1 is the best.

Before the LSTM model training, pre-processing steps were applied all datasets. Punctuation, unnecessary characters, special symbols, and numbers were eliminated from the text for data cleaning. All words were converted to lowercase and removed stop words. In order to obtain meaningful and successful results in the LSTM model, it needs to reduce data complexity so these preprocessing steps are required. However, when the pre-trained models were examined, it has been observed that these models overcome the problem without a preprocessing step such as removing punctuation, stop words, numbers and lowercase. For this reason, pre-processing steps were only applied to the LSTM model before training.

The model cannot understand text representations directly. The data must be represented by a token for the model to understand using a tokenizer. Tokenization methods split the text into small units as words, characters or subwords after that these small units are converted to ids. In this study, LSTM and the pre-trained model used word tokenization and subword tokenization respectively. The word tokenization splits the text by the word of the sentence based on space or delimiter, there is the OOV (Out of Vocabulary) problem, on the contrary, the subword tokenization used in the pre-trained models handle the OOV word. BART tokenizer uses byte-level Byte-Pair-Encoding, while T5 and Pegasus are constructed based on the Sentences Piece subword.

Considering the training time and equipment, the open sources datasets were used whole data as well as different rates to keep the training time short. For all model training, whole resume data was used. In the BART-Large model finetuning, whole data was used from XSUM, News Summary and resume but for Amazon Fine Food Review dataset and CNN/Daily Mail, 20000 training data and 2000 validation data were used. In the LSTM model, 11332 for Xsum dataset, 13368 for CNN/Daily Mail dataset, 10000 for Amazon Review dataset used and additionally, News Summary and Resume datasets are all used. In the PEGASUS and T5 model, 13368 for CNN/Daily Mail dataset, 10000 for Amazon Review dataset used and additionally, Xsum, News Sum and Resume datasets are all used.

*Table 1. Rouge Score for Xsum Dataset*

| Models | XSum ROUGE_1 P | R | F | ROUGE_2 P | R | F | ROUGE_L P | R | F |
|---|---|---|---|---|---|---|---|---|---|
| LSTM | 0,07 | 0,50 | 0,11 | 0,01 | 0,12 | 0,02 | 0,06 | 0,45 | 0,10 |
| bart-base | 0,44 | 0,39 | 0,40 | 0,19 | 0,17 | 0,18 | 0,35 | 0,32 | 0,33 |
| bart-large | **0,45** | **0,44** | **0,44** | **0,23** | **0,22** | **0,22** | **0,38** | **0,37** | **0,37** |
| t5-base | 0,80 | 0,05 | 0,10 | 0,34 | 0,02 | 0,02 | 0,58 | 0,03 | 0,07 |
| pegasus-x-base | 0,81 | 0,05 | 0,10 | 0,36 | 0,02 | 0,04 | 0,59 | 0,04 | 0,07 |

*Table 2. Rouge Score for CNN/Daily Mail Dataset*

| Models | CNN/Daily Mail ROUGE_1 P | R | F | ROUGE_2 P | R | F | ROUGE_L P | R | F |
|---|---|---|---|---|---|---|---|---|---|
| LSTM | 0,13 | 0,76 | 0,26 | 0,05 | 0,42 | 0,09 | 0,12 | 0,73 | 0,21 |
| bart-base | 0,32 | 0,34 | 0,32 | 0,14 | 0,15 | 0,14 | 0,25 | 0,27 | 0,25 |
| bart-large | 0,28 | 0,45 | 0,34 | 0,12 | 0,19 | 0,14 | 0,21 | 0,34 | 0,25 |
| t5-base | 0,98 | 0,09 | 0,16 | 0,85 | 0,07 | 0,14 | 0,88 | 0,08 | 0,14 |
| pegasus-x-base | 0,98 | 0,09 | 0,16 | 0,89 | 0,08 | 0,15 | 0,89 | 0,08 | 0,15 |

*Table 3. Rouge Score for News Summary Dataset*

| Models | News Summary ROUGE_1 P | R | F | ROUGE_2 P | R | F | ROUGE_L P | R | F |
|---|---|---|---|---|---|---|---|---|---|
| LSTM | 0,70 | 0,22 | 0,32 | 0,41 | 0,11 | 0,16 | 0,66 | 0,21 | 0,30 |
| bart-base | 0,42 | 0,43 | 0,42 | 0,20 | 0,21 | 0,20 | 0,30 | 0,31 | 0,31 |
| bart-large | **0,48** | **0,52** | **0,50** | **0,24** | **0,27** | **0,25** | **0,35** | **0,39** | **0,37** |
| t5-base | 0,95 | 0,25 | 0,37 | 0,83 | 0,22 | 0,32 | 0,86 | 0,23 | 0,34 |
| pegasus-x-base | 0,94 | 0,24 | 0,36 | 0,82 | 0,21 | 0,32 | 0,85 | 0,22 | 0,32 |

*Table 4. Rouge Score for Amazon Fine Food Review Dataset*

| Models | Amazon Fine Food Review ROUGE_1 P | R | F | ROUGE_2 P | R | F | ROUGE_L P | R | F |
|---|---|---|---|---|---|---|---|---|---|
| LSTM | 0,03 | 0,27 | 0,05 | 0,01 | 0,07 | 0,01 | 0,03 | 0,26 | 0,05 |
| bart-base | 0,33 | 0,28 | 0,29 | 0,19 | 0,15 | 0,16 | 0,32 | 0,27 | 0,28 |
| bart-large | 0,19 | 0,17 | 0,18 | 0,06 | 0,04 | 0,04 | 0,19 | 0,17 | 0,17 |
| t5-base | 0,80 | 0,05 | 0,10 | 0,42 | 0,03 | 0,05 | 0,77 | 0,05 | 0,09 |
| pegasus-x-base | 0,64 | 0,14 | 0,21 | 0,17 | 0,04 | 0,05 | 0,38 | 0,08 | 0,12 |

*Table 5. Rouge Score for Resume Dataset*

| Models | Resume Dataset ROUGE_1 P | R | F | ROUGE_2 P | R | F | ROUGE_L P | R | F |
|---|---|---|---|---|---|---|---|---|---|
| LSTM | 0,18 | 0,38 | 0,22 | 0,11 | 0,25 | 0,13 | 0,17 | 0,36 | 0,21 |
| bart-base | 0,89 | 0,34 | 0,44 | 0,77 | 0,31 | 0,39 | 0,83 | 0,33 | 0,42 |
| bart-large | **0,90** | **0,53** | **0,64** | **0,78** | **0,46** | **0,56** | **0,86** | **0,51** | **0,62** |
| t5-base | 0,75 | 0,27 | 0,36 | 0,66 | 0,24 | 0,32 | 0,71 | 0,26 | 0,34 |
| pegasus-x-base | 0,65 | 0,23 | 0,32 | 0,56 | 0,19 | 0,25 | 0,62 | 0,19 | 0,30 |

*Table 6. Rouge Score for Resume in Finetuned BART Base Model*

| Models | Resume Dataset ROUGE_1 P | R | F | ROUGE_2 P | R | F | ROUGE_L P | R | F |
|---|---|---|---|---|---|---|---|---|---|
| bart-base | 0,13 | 0,38 | 0,20 | 0,07 | 0,22 | 0,11 | 0,11 | 0,30 | 0,16 |
| Bart-base-resume | 0,89 | 0,34 | 0,44 | 0,77 | 0,31 | 0,39 | 0,83 | 0,33 | 0,42 |
| Bart-base-xsum-resume | 0,93 | 0,41 | 0,52 | 0,83 | 0,37 | 0,46 | 0,86 | 0,40 | 0,50 |
| Bart-base-cnn-resume | 0,82 | 0,43 | 0,53 | 0,75 | 0,39 | 0,48 | 0,77 | 0,42 | 0,51 |
| Bart-base-news summary-resume | **0,83** | **0,45** | **0,54** | **0,75** | **0,41** | **0,49** | **0,77** | **0,43** | **0,52** |
| Bart-base-amazon-resume | 0,77 | 0,48 | 0,57 | 0,66 | 0,42 | 0,50 | 0,69 | 0,41 | 0,51 |

*Table 7. Rouge Score for Resume in Finetuned BART Large Model*

| Models | Resume Dataset | | | | | | | | |
|---|---|---|---|---|---|---|---|---|---|
| | ROUGE_1 | | | ROUGE_2 | | | ROUGE_L | | |
| | P | R | F | P | R | F | P | R | F |
| bart-large | 0,08 | 0,23 | 0,12 | 0,05 | 0,16 | 0,08 | 0,06 | 0,17 | 0,09 |
| Bart-large-resume | 0,90 | 0,53 | 0,64 | 0,78 | 0,46 | 0,56 | 0,86 | 0,51 | 0,62 |
| Bart-large-xsum-resume | 0,68 | 0,48 | 0,54 | 0,57 | 0,48 | 0,44 | 0,66 | 0,47 | 0,52 |
| Bart-large-cnn-resume | 0,76 | 0,45 | 0,54 | 0,66 | 0,40 | 0,48 | 0,73 | 0,44 | 0,53 |
| Bart-large-news summary-resume | 0,60 | 0,44 | 0,49 | 0,46 | 0,33 | 0,37 | 0,57 | 0,42 | 0,46 |
| Bart-large-amazon-resume | 0,71 | 0,53 | 0,59 | 0,56 | 0,42 | 0,47 | 0,67 | 0,51 | 0,56 |

The summarization performance of the LSTM model and fine-tuned pre-trained models were evaluated on the test data. In Table 1, the performance of the LSTM model and the BART, BART-Large, T5 and Pegasus-x models which were trained and finetuned with the Xsum dataset were evaluated on Xsum test data. When the results were analyzed, it was observed that the highest ROUGE was obtained with the BART-Large model. In Table 2, the performance of the models trained and fine-tuned with the CNN/DailyMail dataset is evaluated on the CNN/DailyMail test data. It is seen that the BART-Large model has the highest ROUGE score. Likewise, in Table 3, the performance of the models which were trained and finetuned using News Summary dataset were compared to the News Summary test data. The BART-Large model gives the best ROUGE. After training and finetuning with the Amazon Fine Food Review dataset, which is the last of the open-source datasets we used in the study, the models were tested with the Amazon Fine Food Review test data and the best result was obtained with the Bart-Large (Table 4). The large model did not give better result than BART-Base on CNN/Daily Mail and Amazon Fine Food Review datasets. The reduction in the number of data in order to keep the training period short can be interpreted as the reason for this. Finally, LSTM model and pre-trained models were trained and fine-tuned with Resume dataset, which is our own custom dataset created for this study. When the results are compared in Table 5, it was seen that the Bart-Large model gave the best results for resume summarization.

When the results of all models were compared, it was observed that the BART-Large model has the best performance. The performance of the BART-Large model on the resume dataset was evaluated and focused on increasing BART-Large models, which were finetuned with the open source datasets and the results of which are given in Table1, Table 2, Table 3, Table 4 and Table 5, were finetuned with the resume dataset to increase the performance of the BART-Base and BART-Large model on the resume dataset. It has been observed that the performance of the BART-Base model, which was fine-tuned with open source datasets and then fine-tuned with the resume dataset, it increased on the resume dataset when compared to the model that was fine-tuned with only the resume dataset. Result of models were compared in Table 6. In this table, facebook/bart-base was not finetuned any dataset. BART-Base-Resume model was finetuned only the resume dataset. BART-Base-Xsum-Resume, BART-Base-CNN/DailyMail-Resume, BART-Base-News Summary-Resume and BART-Base-Amazon Fine Food Review-Resume models were finetuned with Xsum, CNN/DailyMail, News Summary and Amazon Fine Food Review dataset respectively, and then fine-tuned with the resume dataset. BART-Base-News Summary-Resume model gave the best ROUGE score on the resume dataset. BART-Large model performance was compared in Table 7. Although the performance of the finetuned models has increased on the resume dataset, they have not achieved the performance of the BART-Large-resume model. BART-Large-resume model gave the best ROUGE score on the resume dataset. This may be because the large model with many parameters and model fine-tuning with open source dataset causing complexity in the model, and it may not have shown an increase in performance with a small number of data.

IV. RESULTS

In this study, the performance of the LSTM and pre-trained models were evaluated on open-source datasets and our resume dataset. For training and finetuning, open source datasets which are Xsum, CNN/DailyMail, News Summary and Amazon Fine Food were used. Apart from these datasets, we created our dataset which is name resume dataset. It has many information such as language, education, experience, personal information, and skills and contains 75 resumes. This study focused on resume text classification. LSTM, pre-trained models and finetuned models were evaluated on resume dataset. BART-Large-resume model that was finetuned with resume dataset gave the best performance. This study can be improved with large resume dataset.

References

[1] Song, Shengli, Haitao Huang, and Tongxiao Ruan. "Abstractive text summarization using LSTM-CNN based deep learning." Multimedia Tools and Applications 78 (2019): 857-875.

[2] Hanunggul, Puruso Muhammad, and Suyanto Suyanto. "The impact of local attention in lstm for abstractive text summarization." 2019 International Seminar on Research of Information Technology and Intelligent Systems (ISRITI). IEEE, 2019.

[3] Zolotareva, Ekaterina, Tsegaye Misikir Tashu, and Tomás Horváth. "Abstractive Text Summarization using Transfer Learning." ITAT. 2020.

[4] Ranganathan, Jaishree, and Gloria Abuka. "Text summarization using transformer model." 2022 Ninth International Conference on Social Networks Analysis, Management and Security (SNAMS). IEEE, 2022

[5] Zhang, Jingqing, et al. "Pegasus: Pre-training with extracted gap-sentences for abstractive summarization." International Conference on Machine Learning. PMLR, 2020.

[6] Lalitha, Evani, et al. "Text Summarization of Medical Documents using Abstractive Techniques." 2023 2nd International Conference on Applied Artificial Intelligence and Computing (ICAAIC). IEEE, 2023.

[7] Bohra, Mrinmoi, Pankaj Dadure, and Partha Pakray. "Comparative analysis of T5 model for abstractive text summarization on different datasets." (2022).

[8] Gupta, Anushka, et al. "Automated news summarization using transformers." Sustainable Advanced Computing: Select Proceedings of ICSAC 2021. Singapore: Springer Singapore, 2022. 249-259.

[9] Yadav, Hemant, Nehal Patel, and Dishank Jani. "Fine-Tuning BART for Abstractive Reviews Summarization." Computational Intelligence: Select Proceedings of InCITe 2022. Singapore: Springer Nature Singapore, 2023. 375-385.

[10] Rehman, T., Das, S., Sanyal, D. K., & Chattopadhyay, S. (2022, August). An Analysis of Abstractive Text Summarization Using Pre-trained Models. In Proceedings of International Conference on Computational Intelligence, Data Science and Cloud Computing: IEM-ICDC 2021 (pp. 253-264).

[11] Narayan, Shashi, Shay B. Cohen, and Mirella Lapata. "Don't give me the details, just the summary! topic-aware convolutional neural networks for extreme summarization." arXiv preprint arXiv:1808.08745 (2018).

[12] Hermann, Karl Moritz, et al. "Teaching machines to read and comprehend." Advances in neural information processing systems 28 (2015).

[13] McAuley, Julian John, and Jure Leskovec. "From amateurs to connoisseurs: modeling the evolution of user expertise through online reviews." Proceedings of the 22nd international conference on World Wide Web. 2013.

[14] Xue, Hao, Du Q. Huynh, and Mark Reynolds. "SS-LSTM: A hierarchical LSTM model for pedestrian trajectory prediction." 2018 IEEE Winter Conference on Applications of Computer Vision (WACV). IEEE, 2018.

[15] Lewis, Mike, et al. "Bart: Denoising sequence-to-sequence pre-training for natural language generation, translation, and comprehension." arXiv preprint arXiv:1910.13461 (2019).

[16] Ahmed, Rafi, et al. "An overview of Pegasus." Proceedings RIDE-IMS93: Third International Workshop on Research Issues in Data Engineering: Interoperability in Multidatabase Systems. IEEE, 1993.

[17] Carmo, Diedre, et al. "Ptt5: Pretraining and validating the t5 model on brazilian portuguese data." arXiv preprint arXiv:2008.09144 (2020).